\documentclass[sn-mathphys-num]{sn-jnl}
\usepackage{graphicx}%
\usepackage{multirow}%
\usepackage{amsmath,amssymb,amsfonts}%
\usepackage{amsthm}%
\usepackage{mathrsfs}%
\usepackage[title]{appendix}%
\usepackage[dvipsnames]{xcolor}
\usepackage{textcomp}%
\usepackage{manyfoot}%
\usepackage{booktabs}%
\usepackage{algorithm}%
\usepackage{algorithmicx}%
\usepackage{algpseudocode}%
\usepackage{listings}%
\usepackage{rotating}
\usepackage{comment}
\usepackage[english]{babel}
\usepackage{subcaption}
\usepackage{array}
\usepackage{lineno}
\usepackage{colortbl}
\usepackage{tikz}
\definecolor{darkgreen}{rgb}{0 0.6 0}

\theoremstyle{thmstyleone}%
\theoremstyle{thmstyletwo}%

\theoremstyle{thmstylethree}%

\usepackage{tocloft}   

\cftpagenumbersoff{figure}

\cftsetindents{figure}{0em}{5em} 
\begin{document}
\title[Article Title]{\centering Systematic Evaluation and Guidelines for Segment Anything Model in Surgical Video Analysis}

\author[1]{\fnm{Cheng} \sur{Yuan}}\email{c.yuan@sjtu.edu.cn}
\equalcont{These authors contributed equally to this work.}

\author[1]{\fnm{Jian} \sur{Jiang}}\email{jian.jiang@sjtu.edu.cn}
\equalcont{These authors contributed equally to this work.}

\author[1]{\fnm{Kunyi} \sur{Yang}}\email{yangkunyi@sjtu.edu.cn}
\equalcont{These authors contributed equally to this work.}

\author[1]{\fnm{Lv} \sur{Wu}}
\author[1]{\fnm{Rui} \sur{Wang}}
\author[1]{\fnm{Zi} \sur{Meng}}
\author[1]{\fnm{Haonan} \sur{Ping}}
\author[1]{\fnm{Ziyu} \sur{Xu}}
\author[1]{\fnm{Yifan} \sur{Zhou}}
\author[1]{\fnm{Wanli} \sur{Song}}
\author[2]{\fnm{Hesheng} \sur{Wang}}
\author[3]{\fnm{Yueming} \sur{Jin}}
\author[4]{\fnm{Qi} \sur{Dou}}
\author*[1]{\fnm{Yutong} \sur{Ban}}\email{yban@sjtu.edu.cn}

\affil[1]{\orgdiv{Global College}, \orgname{Shanghai Jiao Tong University}, \city{Shanghai},\country{China}}
\affil[2]{\orgdiv{Department of Automation}, \orgname{Shanghai Jiao Tong University}, \city{Shanghai}, \country{China}}
\affil[3]{\orgdiv{Department of Biomedical Engineering and Department of Electrical and Computer Engineering}, \orgname{National University of Singapore}, \country{Singapore}}
\affil[4]{\orgdiv{Department of Computer Science and Engineering}, \orgname{The Chinese University of Hong Kong}, \state{Hong Kong SAR}, \country{China}}


\abstract{Surgical video segmentation is critical for AI to interpret spatial-temporal dynamics in surgery, yet model performance is constrained by limited annotated data. The SAM2 model, pretrained on natural videos, offers potential for zero-shot surgical segmentation, but its applicability in complex surgical environments, with challenges like tissue deformation and instrument variability, remains unexplored. We present the first comprehensive evaluation of the zero-shot capability of SAM2 in 9 surgical datasets (17 surgery types), covering laparoscopic, endoscopic, and robotic procedures. We analyze various prompting (points, boxes, mask) and {finetuning (dense, sparse) strategies}, robustness to surgical challenges, and generalization across procedures and anatomies. Key findings reveal that while SAM2 demonstrates notable zero-shot adaptability in structured scenarios (e.g., instrument segmentation, {multi-organ segmentation}, and scene segmentation), its performance varies under dynamic surgical conditions, highlighting gaps in handling temporal coherence and domain-specific artifacts. These results highlight future pathways to adaptive data-efficient solutions for the surgical data science field.}




\maketitle

\section{Introduction}\label{sec1}
The rapid development of computer vision has seen foundation models demonstrating impressive zero-shot and few-shot capabilities across various tasks. {The Segment Anything Model (SAM) exemplifies this trend, leveraging large-scale datasets to learn highly generalizable object representations. SAM operates as a class-agnostic segmentation model, producing instance masks for individual objects based on user prompts.Its formulation as a promptable, class-agnostic segmentation foundation model enables training on diverse and often incompletely labeled datasets, facilitating broad generalization. SAM demonstrated significant potential across numerous domains as the first foundation model released for this class-agnostic segmentation paradigm.} However, when segmenting video data, {it requires explicit prompts} for each frame, which can be time-consuming and impractical for dynamic scenes.

\begin{figure*}[!t]
    \begin{center}
    \includegraphics[width=1.0\linewidth]{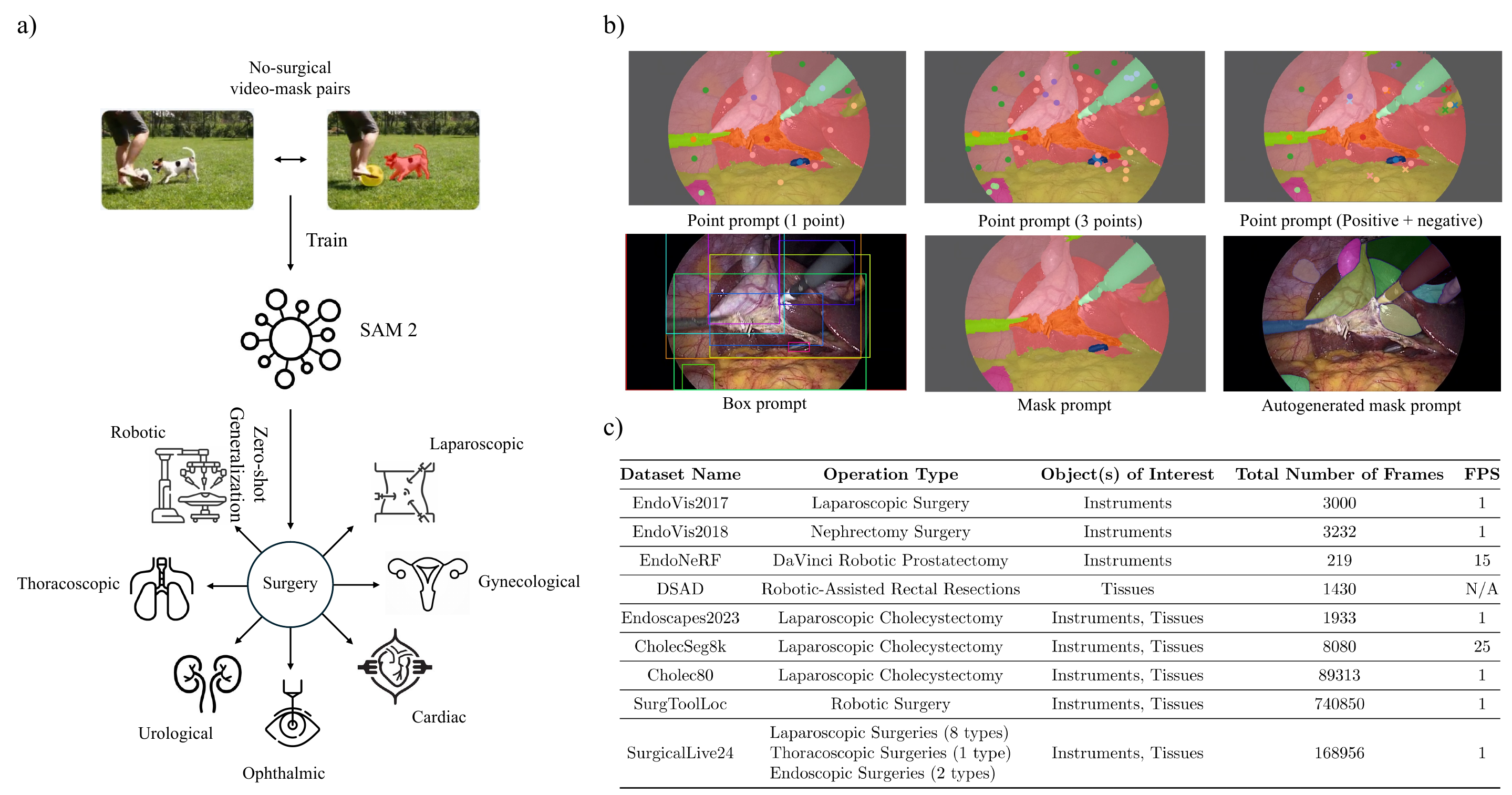}
    \caption{SAM2 overview and evaluated datasets. 
a) SAM2 application across different surgical specialties;
b) Quantitative segmentation performance across SAM2 prompt types, including point, box and mask prompts; 
c) Datasets used for SAM2 evaluation, including 8 public datasets and 1 private dataset.}
    \label{Fig:merged_1}
    \end{center}
\end{figure*}

Recently, Segment Anything Model 2 (SAM2) has extended the zero-shot segmentation capabilities of the original SAM to video data. Trained on the SA-V dataset, {comprising} 35.5 million masks across 50.9 thousand videos, SAM2 demonstrates {strong zero-shot performance} for video segmentation. {A key advancement in SAM2 is the incorporation of a memory bank, enabling the propagation of prompts from an initial frame throughout the video sequence.} This feature {} makes it particularly well-suited for the surgery scene segmentation and tracking of surgical tools in surgical videos. As shown in Figure~\ref{Fig:merged_1}a), SAM2 {demonstrates applicability across diverse surgical assistant tasks, including} robotic assistance and video processing in laparoscopic, gynecological, cardiac, thoracoscopic, ophthalmic, and urological surgeries.
SAM2 supports a versatile range of prompting mechanisms, {allowing users to interactively specify objects of interest} across video frames using points, bounding boxes, or pixel masks, as shown in Figure~\ref{Fig:merged_1}b). Each prompt is associated with additional metadata to provide context: an object ID indicating the target instance and a frame index specifying its temporal location.  This flexible input format enables precise guidance for consistent object segmentation across multiple frames. This work systematically assessed SAM2's performance across diverse surgical datasets, covering laparoscopic instrument segmentation, scene understanding, and tool detection tasks. An overview of the characteristics (including type, number, and fps) of all the used datasets can be found in Figure~\ref{Fig:merged_1}c).

{To contextualize such advancements of SAM2 within the broader evolution of surgical analysis, we further review the significant progresses achieved by related conventional methods in both temporal tasks (such as phase recognition) and spatial tasks (like segmentation and detection). As below, related works are introduced in three parts: surgical temporal task, surgical spatial task, and foundation model in segmentation.}

{Surgical procedures consist of sequential, well-defined steps performed continuously without interruption. This temporal structure enables the decomposition of surgeries into discrete process units known as surgical phases or steps. Consequently, surgical phase recognition represents a fundamental temporal task in surgical video analysis.} The first works on the analysis of surgical procedures by decomposing them into subparts were published in 2001 by Jannin et al.~\cite{jannin2003model} and MacKenzie et al.~\cite{mackenzie2001hierarchical}. Due to the tremendous classification and recognition capabilities of multilayer neural networks, deep learning models are used progressively in surgical recognition tasks, and their potential is confirmed by many studies. In early studies, Convolutional Neural Networks (CNNs) were shown to be more effective in feature extraction than state-of-the-art statistical methods. One strategy for surgical workflow analysis is to perform a frame-wise classification directly based on extracted features from a single image. Zhang et al.~\cite{zhang2021surgical} improved this approach using a 3D CNN to extract features. Pradeep and Sinha~\cite{pradeep2021spatio} extracted spatio-temporal features from 64 images together and used them for classification directly. Another strategy is based on the recurrent neural network (RNN) for sequential data. Zia et al.~\cite{zia2018surgical} achieved the best results in their study with bidirectional single-layer RNN. 

In the field of spatial feature perception, instrument and anatomy segmentation is critical for surgical scene understanding, as it enables the development of computer-assisted systems for the tracking of regions of interest (RoIs)~\cite{chmarra2007systems}, pose estimation~\cite{du2018articulated}, and surgical phase estimation~\cite{sanchez2022data}. Considerable research effort~\cite{iglovikov2018ternausnet,jin2019incorporating,zhao2020learning,meinhardt2022trackformer,gonzalez2020isinet,zhao2022trasetr,baby2023forks,sheng2024surgical,dlv3p, selfpsam, surgicalsam, persam, medsam, murali2024cyclesam} has been made in addressing this problem. CNN-based models have been common for surgical RoI segmentation. Initial efforts used Fully Convolutional Networks (FCN) for instrument parts segmentation. Later, the 2017 Endoscopic Vision Robotic Instrument Segmentation Challenge (EndoVis2017) and its 2018 version (EndoVis2018) introduced the instrument subtype segmentation task. Thus, most methods adapted FCN-based models to perform semantic segmentation of the different instruments present on each frame. Some models leverage additional priors like optical and motion flow, stereoscopic information, or saliency maps. More recent approaches have modified the original task by including weak supervision~\cite{zhao2020learning}, domain adaptation~\cite{zhao2021one}, and image generation~\cite{colleoni2021robotic}.

A foundation model is any model that is trained on broad data that can be fine-tuned to a wide range of downstream tasks. Image segmentation foundation models have revolutionized the field of image segmentation, demonstrating wide generalizability and impressive performance by training on massive amounts of data to learn general representations. Prompt engineering further improves the segmentation capability of these models. Given appropriate prompts as additional input, these models can handle various zero-shot tasks across domains and produce reliable segmentation during inference. Unlike these broad successes, medical image segmentation is often limited by issues such as expensive data acquisition and time-consuming annotation processing, resulting in a lack of massive public datasets available for training. Thus, it is desirable to leverage transfer learning from the natural image domain for robust medical image segmentation.

 {This study includes three key contributions: (1) We develop a unified framework for multi-object segmentation in surgical videos, encompassing surgical instruments, organs, and fine-grained tissue structures. (2) We conduct a large-scale empirical evaluation of zero-shot segmentation in surgery, comparing prompting strategies, re-initialization methods, and an automatic segmentation pipeline across 9 datasets covering 17 surgical procedure types. (3) We present an in-depth study of SAM2 training strategies for surgical scene segmentation under diverse configurations, yielding actionable guidelines for deploying SAM2 in surgical video analysis.}

\section{Results}\label{sec2}

{Overall performance comparisons on five annotated datasets are shown in Table~\ref{tab:sam2_merged_results}. Here, 5 kinds of prompt types are used for evaluating prompting impacts, that are 1 center point, 1 random point, 3 random points, box, and mask. In addition, in every prompting experiment, 3 kinds of reinitialization strategies are set to find the changes of tracking performance, which are based on no reinitialization, 1 reinitialization per 30 frames, and 1 reinitialization per 60 frames. This reinitialization strategy involves resetting the prompt with the ground truth annotation of the current frame and clearing the model's memory to prevent error accumulation. To understand our finetuning approach, it is important to note that the SAM2 architecture consists of an image encoder, a prompt encoder, a mask decoder, and a memory module (comprising a memory encoder and memory attention mechanism) for temporal propagation. The Finetuned SAM2 Variants section of the table details our ablation studies on adapting the model. This finetuning is performed using an image-based, dense strategy, meaning the model is trained on every frame with valid annotations. For these experiments, we use the setup without re-initialization to evaluate the direct impact of finetuning on the model's core components.}

{Specifically, point-based prompts exhibit significant variability. Mask prompting achieves the highest overall segmentation performance, followed by bounding boxes. Three random points substantially outperform single points across most datasets. In addition, our proposed reinitialization strategy consistently enhances surgical video segmentation performance across all included datasets. Reinitialization at 30-frame intervals yields superior gains compared to reinitialization at 60-frame intervals. For finetuning ablation results, combined with the mask prompting way, training the mask decoder and the prompt encoder, meanwhile freezing others gets the best results in almost all datasets, which proposes an effective finetuning strategy. Finetuning the image encoder causes a decrease in segmentation and recognition accuracy, possibly because the limited surgical data cannot afford the training of the image encoder with huge parameters.}

\begin{table*}[!htb]
\centering
\caption{Performance comparison of SAM-based methods across annotated surgical datasets (numbers in \%). Dataset-level metrics are averaged across all subcategories. For EndoNeRF, values are averaged between Cutting and Pulling subsets. MD: Training Mask Decoder only; MD+PE: Training Mask Decoder and Prompt Encoder; MD+PE+IE: Training Mask Decoder, Prompt Encoder and Image Encoder.}
\resizebox{\textwidth}{!}{
\begin{tabular}{l|cc|cc|cc|cc|ccc|ccc}
\toprule
\hiderowcolors  
\multirow{2}{*}{\textbf{Method}} & \multicolumn{2}{c|}{\textbf{EndoVis2017}~\cite{endovis17}} & \multicolumn{2}{c|}{\textbf{EndoVis2018}~\cite{endovis18}} & \multicolumn{2}{c|}{\textbf{EndoNeRF}~\cite{endonerf}} & \multicolumn{2}{c|}{\textbf{DSAD}~\cite{dsad}} & \multicolumn{3}{c|}{\textbf{Endoscapes2023~\cite{murali2023endoscapes}}} & \multicolumn{3}{c}{\textbf{CholecSeg8k}~\cite{hong2020cholecseg8k}}\\
\cmidrule{2-15}
 & mIoU & mDice & mIoU & mDice & mIoU & mDice & mIoU & mDice & mIoU & mDice & mAP & mIoU & mDice & mAP \\
\midrule
\multicolumn{15}{c}{\textbf{Compared Methods}} \\
\midrule
Surgical-DeSAM~\cite{sheng2024desam} & 82.41 & 89.62 & 84.91 & 90.70 & - & - & - & - & - & - & - & - & - & - \\
SelfPromptSAM~\cite{wu2023selfpromptsam} & - & - & - & - & - & - & - & - & - & - & 1.70 & - & - & 1.90 \\
SurgicalSAM~\cite{yue2024surgicalsam} & - & - & - & - & - & - & - & - & - & - & 5.70 & - & - & 8.90 \\
PerSAM~\cite{zhang2305persam} & - & - & - & - & - & - & - & - & - & - & 1.30 & - & - & 2.90 \\
GF-SAM~\cite{zhang2024gfsam} & - & - & - & - & - & - & - & - & - & - & 2.60 & - & - & 6.30 \\
CycleSAM~\cite{murali2024cyclesam} & - & - & - & - & - & - & - & - & - & - & 15.90 & - & - & 22.40 \\
\midrule
\rowcolor{white}\multicolumn{15}{c}{\textbf{Vanilla SAM2 with Different Prompts}} \\
\midrule
SAM2-1Point-Center  & 46.67 & 50.81 & 50.80 & - & 89.41 & 93.88 & 15.70 & 17.72 & 20.79 & 26.07 & 16.94 & 61.98 & 70.08 & 54.90 \\
SAM2-1Point-Center-Reinit 60 & 51.12 & 53.71 & - & - & 90.21 & 94.20 & 20.03 & 22.52 & 24.76 & 30.60 & 22.60 & 62.08 & 70.20 & 54.02 \\
SAM2-1Point-Center-Reinit 30 & 65.01 & 70.30 & 53.96 & - & 90.21 & 94.31 & 26.21 & 29.26 & 37.84 & 45.49 & 37.84 & 62.38 & 70.40 & 54.30 \\
SAM2-1Point  & 43.65 & 47.51 & 40.75 & - & 89.08 & 93.84 & 14.78 & 16.71 & 22.06 & 27.48 & 16.74 & 63.43 & 71.53 & 52.94 \\
SAM2-1Point-Random-Reinit 60 & 47.09 & 50.81 & 55.54 & - & 90.18 & 94.18 & 19.10 & 21.58 & 23.08 & 28.94 & 16.53 & 65.09 & 73.32 & 46.99 \\
SAM2-1Point-Random-Reinit 30 & 62.51 & 68.20 & 50.09 & - & 85.20 & 87.52 & 25.43 & 28.62 & 33.78 & 41.12 & 22.85 & 64.72 & 73.06 & 47.08 \\
SAM2-3Points-Random & 49.71 & 53.34 & 55.48 & - & 89.49 & 94.49 & 17.48 & 19.63 & 27.30 & 33.44 & 19.46 & 71.24 & 79.17 & 59.41 \\
SAM2-3Points-Random-Reinit 60 & 63.05 & 67.58 & 50.69 & - & 89.50 & 93.70 & 21.48 & 24.15 & 29.37 & 36.16 & 23.10 & 71.64 & 79.50 & 54.64 \\
SAM2-3Points-Random-Reinit 30 & 69.53 & 74.56 & 48.35 & - & 85.01 & 87.75 & 27.57 & 30.81 & 41.23 & 49.09 & 31.81 & 72.58 & 80.39 & 59.72 \\
SAM2-Bbox & 50.44 & 53.67 & 60.52 & - & 90.28 & 94.34 & 18.76 & 20.96 & 38.94 & 44.88 & 54.97 & 83.34 & 89.00 & 87.39 \\
SAM2-Bbox-Reinit 60 & 65.21 & 69.28 & 62.70 & - & 90.29 & 94.43 & 23.24 & 25.87 & 49.32 & 56.39 & 64.62 & 84.31 & 89.89 & 88.61 \\
SAM2-Bbox-Reinit 30 & 73.27 & 77.29 & 69.17 & - & 90.32 & 94.36 & 30.75 & 33.98 & 76.82 & 85.87 & 88.91 & 84.93 & 90.49 & 90.22 \\
SAM2-Mask & 52.63 & 55.52 & 57.68 & - & 84.82 & 88.74 & 19.58 & 21.59 & 47.82 & 51.81 & 39.89 & 88.95 & 92.73 & 94.32 \\
SAM2-Mask-Reinit 60 & 68.55 & 71.91 & 67.58 & - & 85.90 & 89.50 & 24.17 & 26.52 & 63.79 & 66.54 & 64.62 & 90.14 & 93.69 & 95.17 \\
SAM2-Mask-Reinit 30 & 76.15 & 79.31 & 73.06 & - & 85.76 & 89.26 & 32.40 & 35.23 & 98.91 & 99.45 & 100.0 & 91.14 & 94.48 & 94.96 \\
\midrule
\rowcolor{white}\multicolumn{15}{c}{\textbf{Finetuned SAM2 Variants}} \\
\midrule
SAM2-FT-1Point-MD & 73.72 & 78.79 & 46.38 & 50.61 & 80.86 & 88.87 & 63.36 & 72.33 & 30.57 & 36.10 & 42.19 & 67.74 & 75.18 & 74.99 \\
SAM2-FT-1Point-MD+PE & 78.47 & 83.13 & 44.95 & 49.00 & 80.85 & 88.87 & 62.72 & 71.86 & 31.74 & 37.60 & 41.88 & 67.33 & 75.13 & 78.94 \\
SAM2-FT-1Point-MD+PE+IE & 73.98 & 80.01 & 38.86 & 42.77 & 84.18 & 90.82 & 62.12 & 71.45 & 35.96 & 41.89 & 44.33 & 68.87 & 75.99 & 76.40 \\
SAM2-FT-Bbox-MD & 80.49 & 85.19 & 50.46 & 53.63 & 79.97 & 88.37 & 73.62 & 81.29 & 42.51 & 47.96 & 63.16 & 76.14 & 83.09 & 88.08 \\
SAM2-FT-Bbox-MD+PE & 80.08 & 84.83 & 48.86 & 51.91 & 79.99 & 88.38 & 73.94 & 81.45 & 42.05 & 47.47 & 63.56 & 75.88 & 82.86 & 87.89 \\
SAM2-FT-Bbox-MD+PE+IE & 70.91 & 74.53 & 41.73 & 46.23 & 84.51 & 91.05 & 70.54 & 78.77 & 44.64 & 50.89 & 62.19 & 77.98 & 84.86 & 91.90 \\
SAM2-FT-Mask-MD & 83.85 & 87.53 & 50.06 & 52.95 & 85.95 & 92.06 & 80.85 & 86.31 & 51.08 & 53.55 & 70.60 & 81.08 & 86.39 & 93.13 \\
SAM2-FT-Mask-MD+PE & 81.88 & 85.78 & 52.04 & 55.10 & 85.99 & 92.09 & 80.87 & 86.25 & 51.07 & 53.82 & 59.89 & 81.12 & 86.43 & 93.28 \\
SAM2-FT-Mask-MD+PE+IE & 54.06 & 60.55 & 28.53 & 32.75 & 88.97 & 93.91 & 61.30 & 69.99 & 47.99 & 50.68 & 64.26 & 77.36 & 83.60 & 91.23 \\
\bottomrule
\end{tabular}
}
\label{tab:sam2_merged_results}
\end{table*}

\subsection{Instrument Segmentation}\label{subsec2}

\begin{figure*}[hb!]
\centering
\includegraphics[width=1\linewidth]{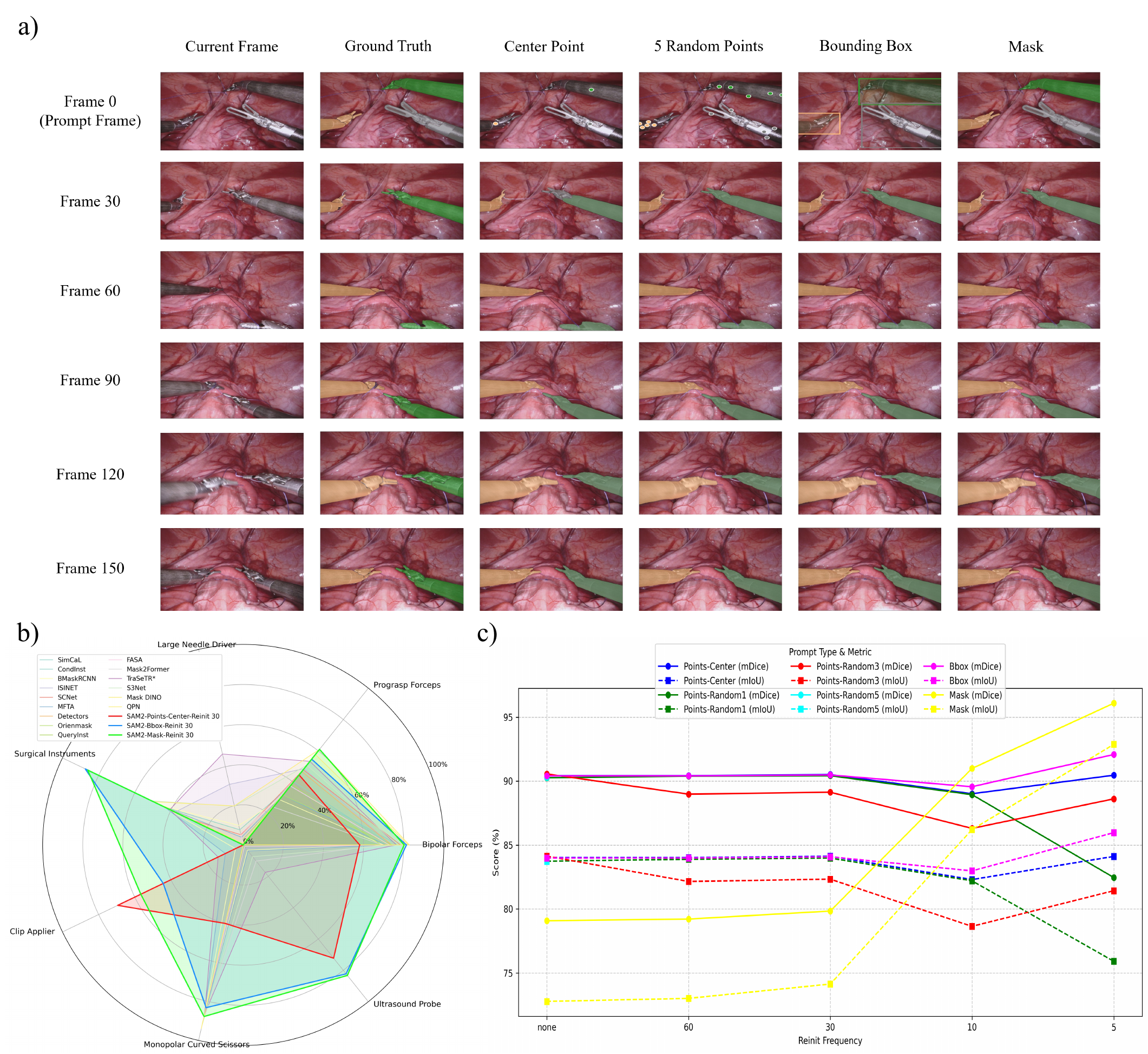}
\caption{Instrument segmentation performance of SAM2 across surgical datasets. 
a) Qualitative comparison showing segmentation results on EndoVis2017 dataset across multiple frames, demonstrating temporal consistency of instrument tracking. Different colors represent distinct surgical instruments identified in each frame;
b) Radar chart displaying class-wise Intersection over Union (IoU) performance for 
various surgical instrument types on the EndoVis2018 dataset. Each axis represents 
a different instrument class with values ranging from 0 to 1, where higher values 
indicate better segmentation accuracy;
c) Quantitative evaluation on the EndoNeRF Cutting dataset showing mean IoU (mIoU) 
and mean Dice coefficient (mDice) metrics. Bar heights represent segmentation accuracy with error bars indicating standard deviation across test samples.}
\label{Fig:merged_2}
\end{figure*}

We use the SAM2 model to analyze three datasets used for surgical tool tracking: EndoVis2017~\cite{endovis17}, EndoVis2018~\cite{endovis18}, and EndoNeRF~\cite{endonerf}. Various prompting strategies were evaluated, including point prompts (center and random), bounding box prompts, and mask prompts. We also investigated the impact of re-initialization at different intervals (30 and 60 frames). {A visualization of the qualitative segmentation results on EndoVis2017 is presented in Figure~\ref{Fig:merged_2}a), showing how different prompts track instruments over time.}

For EndoVis2017, as shown in Table~\ref{tab:sam2_merged_results}, while base performance without re-initialization was modest (mask prompts: 52.63\% mIoU, 55.52\% mDice), implementing 30-frame re-initialization substantially improved results (mask prompts: 76.15\% mIoU, 79.31\% mDice). These results showed a significant improvement over earlier methods which are presented in Supplementary Table 1 such as TernausNet (35.27\% mIoU) and MF-TAPNet (37.35\% mIoU), though not surpassing Surgical-DeSAM's performance (82.41\% mIoU, 89.62\% mDice).

On the EndoVis2018 dataset, as detailed in the per-class radar chart in Figure~\ref{Fig:merged_2}b) and Supplementary Table 2, our SAM2-based method with mask prompts achieved strong performance in segmenting instrument classes, outperforming several state-of-the-art methods. For instance, SAM2-Mask-Reinit 30 achieved an IoU of 60.99\% for Prograsp Forceps and 87.75\% for Monopolar Curved Scissors, compared to QPN’s 60.94\% and 93.93\%, Mask DINO’s 57.67\% and 90.73\%, and TraSeTR’s 53.30\% and 86.30\%, respectively. Notably, SAM2 excelled in challenging classes, such as the Clip Applier with an IoU of 56.66\% versus QPN’s 0.00\%. Other baselines like CondInst (7.77\% for Large Needle Driver, 0.00\% for Clip Applier) and ISINET (30.98\% and 0.00\%) lagged behind in these categories, highlighting SAM2’s robustness across diverse instruments.

On the EndoNeRF dataset, {whose performance is detailed for the Cutting video in Figure~\ref{Fig:merged_2}c), results varied according to the reinit-frequency. According to Table~\ref{tab:sam2_merged_results}, which reports values averaged between the Cutting and Pulling subsets, the vanilla SAM2 with a mask prompt achieves 84.82\% mIoU. With 30-frame re-initialization, this improves to 85.76\% mIoU}.

Re-initialization emerged as a crucial factor across all datasets. In EndoVis2017, it improved the point prompt (1 center point) to 65.01\% mIoU and the bounding box prompt to 73.27\% mIoU. For EndoVis2018, the {30-frame re-initialization improved} box labels from {60. 52\% to 69. 17\% mIoU. The impact of re-initialization is further highlighted in Figure~\ref{Fig:merged_2}c), where more frequent updates boost performance for the EndoNeRF dataset in majority of cases.}

The effectiveness of different prompt types varied across datasets and scenarios. Although mask prompts generally showed superior performance ({EndoVis2017 with 30-frame re-init: 76.15\% mIoU; EndoVis2018 with 30-frame re-init: 73.06\% mIoU}), bounding box prompts were also highly effective ({EndoVis2017 with 30-frame re-init: 73.27\% mIoU; EndoVis2018 with 30-frame re-init: 69.17\% mIoU}). {As expected, point prompts were less robust, with the 3-random-points prompt achieving 69.53\% mIoU on EndoVis2017 and only 48.35\% mIoU on EndoVis2018, both with 30-frame re-initialization.}

\begin{figure*}[htb!]
\centering
\includegraphics[width=1\linewidth]{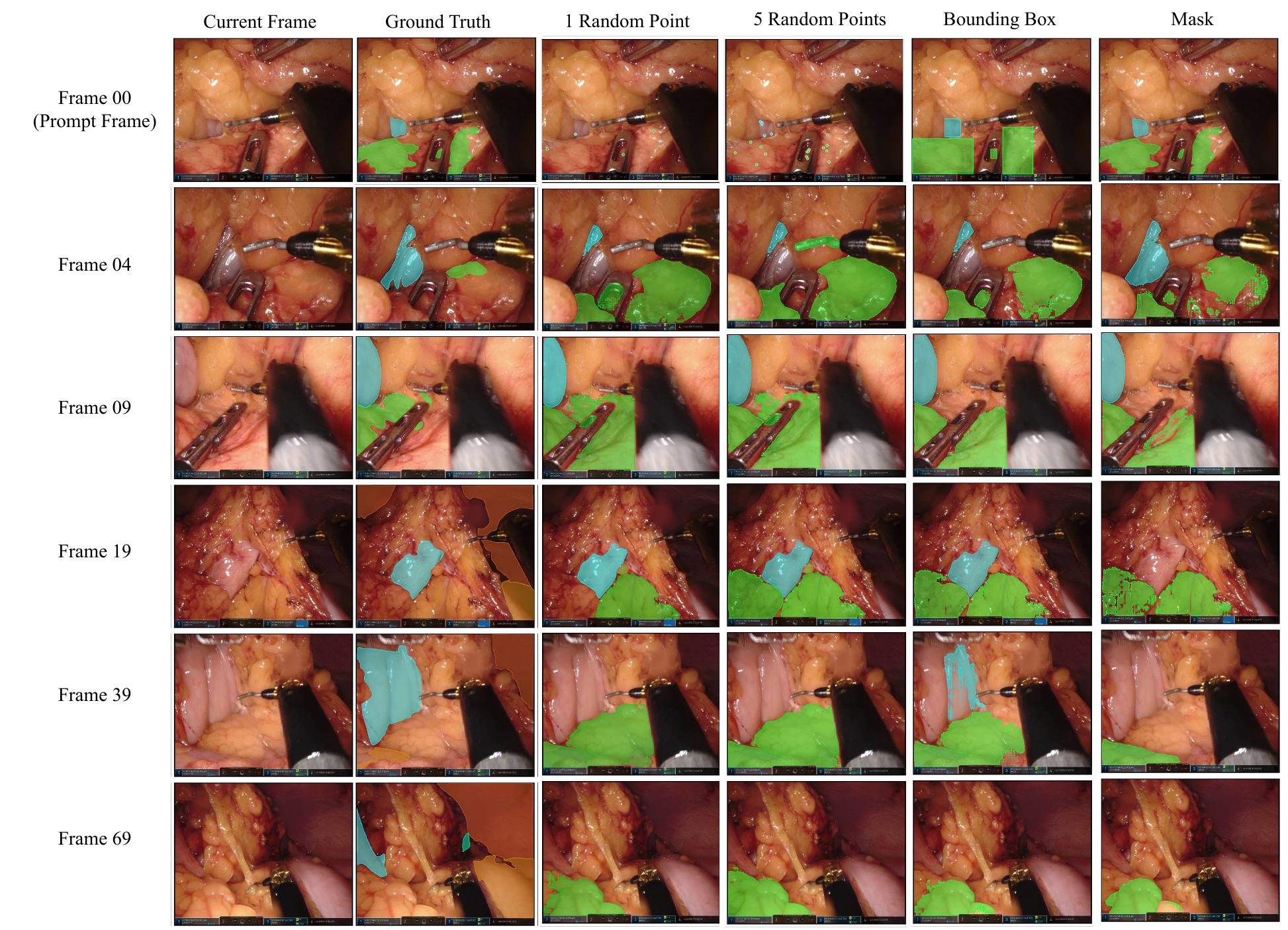}
\caption{Qualitative segmentation results for multi-organ surgical scenes on the DSAD dataset. Each row shows frames from a surgical video sequence, starting with the first row that contains the prompt frame (Frame 00), where colored dots indicate point prompts, colored rectangles indicate box prompts, and colored irregular shapes indicate mask prompts. The following rows show subsequent frames (Frames 04, 09, 19, 39, and 69) with results predicted by SAM2 using the corresponding prompts. Columns represent: 1) Current Frame: the original frame; 2) Ground Truth: manual multi-organ annotations; 3) 1 Random Point: segmentation with a single random point prompt per target; 4) 5 Random Points: segmentation with five random point prompts per target; 5) Bounding Box: segmentation guided by a bounding box for each target; and 6) Mask: segmentation initialized from a predefined mask for each target. The comparison demonstrates the impact of different prompting strategies on multi-organ segmentation quality and temporal consistency across frames.}
\label{dsad}
\end{figure*}

{\subsection{Multi-Organ Segmentation}
We evaluate the effectiveness of SAM2 model on the Dresden Surgical Anatomy Dataset~\cite{dsad} with multi-organ annotations. The same prompting and re-initialization strategies {were used}. As shown in Table~\ref{tab:sam2_merged_results}, mask prompting achieves the best organ segmentation performance, based on not only vanilla SAM2 (32.40\% mIoU and 35.23\% mDice) but also {finetuned} SAM2 (80.87\% mIoU and 86.31\% mDice). For image-based dense finetuning, combined with the mask prompting way, only training the mask decoder achieves the best mDice. Meanwhile, training the mask decoder and the prompt encoder achieves the best mIoU. The visualization of the qualitative segmentation results on the multi-label subset is presented in Figure~\ref{dsad}. Mask prompting perceives more delicate tubular structures and exhibits {fewer false negatives}.}

\subsection{Joint Instrument and Tissue Segmentation}


\begin{figure*}[!ht]
\centering
\includegraphics[width=1\linewidth]{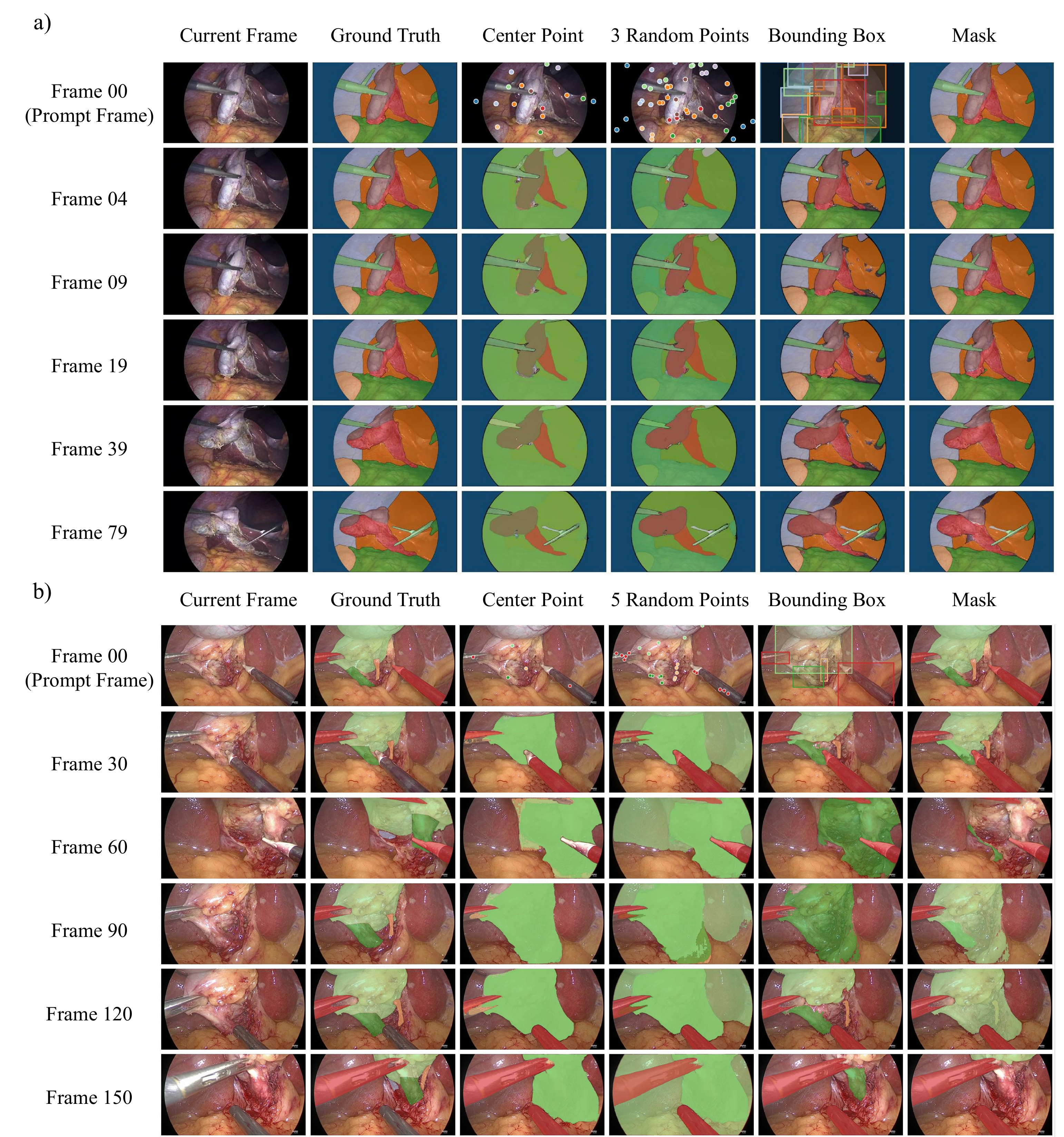}
\caption{Qualitative comparison showing segmentation results on CholecSeg8k and Endoscapes2023 dataset. Mask prompting achieves the best segmentation performance on both datasets. a)  On CholecSeg8k dataset, different colors represent multiple instruments, organs, and tissue. Point prompting cannot recognize the complex background with various tissue; b) On Endoscapes2023 dataset, tubular anatomy structure such as the cystic artery can be recognized more and more better with the expansion of prompting scope.}
\label{seg-50-table}
\end{figure*}

As for instance segmentation, the dataset named Endoscapes2023 is included here. Figure~\ref{seg-50-table} shows the visualization performance of different prompting inferences. As indicated in Table~\ref{tab:sam2_merged_results}, using full masks as guides worked best for image segmentation, better than using boxes or points. The model easily tracked the tools, even when they moved out of and back into view. However, it struggled to find organs or tissues once they were out of view. It’s worth noting that even with slightly inaccurate masks or boxes, SAM2 still did well, showing that it can handle some errors. The results show that detailed shape information is key for good segmentation, especially for organs and tissues, and that using masks or bounding boxes as guides is a promising approach.

As for semantic segmentation, the dataset named CholecSeg8k is included in this study. SAM2 achieved generally best performance on both instruments and tissue appeared in the whole surgery field. Using SAM2 with bounding box prompting or mask prompting can have a much better segmentation outcome compared to {other SAM-based methods; for example, SAM2 with a mask prompt without re-initialization achieves 94.32\% mAP, significantly outperforming CycleSAM's 22.40\% mAP as shown in Table~\ref{tab:sam2_merged_results}}. Furthermore, we assessed the impact of re-initialization at the interval of 60 and 30 frames for each method. Additionally, we tested mask and bounding box prompts with slight positional inaccuracies as shown in Supplementary Section 3. The results indicate that SAM2 maintains strong segmentation performance even with these imperfect prompts, demonstrating its robustness.

The effect of re-initialization varied across the different prompting strategies, as shown in Table~\ref{tab:sam2_merged_results}. In general, introducing periodic re-initialization significantly improves SAM2's performance with mask and bounding box prompting strategies. {On the Endoscapes2023 dataset, 30-frame re-initialization with bounding box prompts leads to a dramatic improvement in mIoU from 38.94\% to 76.82\%. Similarly, mask prompts with 30-frame re-initialization achieve an outstanding mIoU of 98.91\%, indicating the effectiveness of updated information for tracking in dynamic scenes. On the CholecSeg8k dataset, the effect of re-initialization is also evident, though with different magnitudes of improvement. With bounding box prompts, introducing a 30-frame re-initialization modestly increases the mIoU from 83.34\% to 84.93\%. Similarly, for mask prompts, the 30-frame re-initialization results in an improvement in mIoU from 88.95\% to 91.14\%, demonstrating the benefit of providing updated information for maintaining tracking accuracy.}

\subsection{Unlabeled Surgical Video Segmentation}
This section evaluates SAM2 for surgical tool and tissue segmentation in unlabeled videos using three datasets: Cholec80~\cite{twinanda2016endonet}, SurgToolLoc~\cite{zia2023surgical} and SurgicalLive24. For those datasets with no ground truth, we typically start by automatically generating a mask for the first frame with the SAM2 automatic mask generator to identify the object we need to track. The mask and the first frame will then be used to initialize the SAM2 video predictor.


For Cholec80, visualization results are presented in Supplementary Figure 5c) are performed separately with point and mask prompts, where points and mask are generated by the SAM2 automatic mask generator. Successful cases were characterized by clear initial frames containing surgical tools, enabling high-quality mask generation and subsequent accurate tracking.

For SurgToolLoc, the first column of Supplementary Figure 5b) depicts a representative example of the automatically generated mask of the first frame on the SurgToolLoc dataset. The points generated along with the mask by the automatic mask generator are then used as point prompts for the SAM2 video predictor to generate masks in the subsequent frames of the video clip. The qualitative segmentation performance of SAM2 across time frames is demonstrated in the second to fifth column of Supplementary Figure 5b) respectively. SAM2 delivers reliable segmentation results for both surgical tools and tissue when the endoscopic scene is well lit and the surgical tools exhibit smooth motion.

For SurgicalLive24, performance varies depending on the type of surgery (Supplementary Figure 5a)). In narrow surgical video scenarios, such as ureteroscopy (Supplementary Figure 5a)-i)), the overall segmentation effect is not satisfactory. This may be due to the low resolution of the video itself and the blur caused by the rapid changes of objects in narrow scenes. In contrast, in more open and well-lit procedures like laparoscopy, SAM2 can achieve good segmentation and tracking results for surgical instruments. For tissues and organs, it is difficult to track them over long video sequences, due to the unclear boundaries of organs and tissues. Changes in the scene, such as the movement of camera or the introduction of new organs and tissues, can hinder the tracking of tissues and organs.

\begin{table*}[!htb]
\centering
\caption{Performance comparison of Finetuned SAM2 Variants across annotated surgical datasets (numbers in \%). The top section shows an image-based approach, training only on sparsely annotated frames (stride of 4). The bottom section shows a video-based approach, processing all frames but only applying supervision on the sparsely annotated frames. MD: Training Mask Decoder only; MD+PE: Training Mask Decoder and Prompt Encoder; MD+PE+IE: Training Mask Decoder, Prompt Encoder and Image Encoder.}
\resizebox{0.95\textwidth}{!}{ 
\begin{tabular}{l|cc|cc|cc}
\toprule
\hiderowcolors
\multirow{2}{*}{\textbf{Method}} & \multicolumn{2}{c|}{\textbf{EndoVis2017}} & \multicolumn{2}{c|}{\textbf{EndoVis2018}} & \multicolumn{2}{c}{\textbf{CholecSeg8k}}\\
\cmidrule{2-7}
 & mIoU & mDice & mIoU & mDice & mIoU & mDice \\
\midrule
\multicolumn{7}{c}{\textbf{Finetuned SAM2 Variants (Image-based, Sparse Annotation)}} \\
\midrule
SAM2-FT-Point-MD         & 73.89 & 78.74 & 35.54 & 38.42 & 67.84 & 76.14 \\
SAM2-FT-Point-MD+PE      & 74.11 & 79.20 & 45.42 & 49.69 & 68.60 & 76.79 \\
SAM2-FT-Bbox-MD          & 80.82 & 85.36 & 49.55 & 52.69 & 76.10 & 82.95 \\
SAM2-FT-Bbox-MD+PE       & 80.56 & 85.19 & 49.92 & 53.08 & 75.60 & 82.62 \\
SAM2-FT-Mask-MD          & 82.13 & 86.05 & 51.88 & 55.01 & 81.06 & 86.38 \\
SAM2-FT-Mask-MD+PE       & 81.78 & 85.74 & 52.26 & 55.41 & 80.89 & 86.27 \\
\midrule
\rowcolor{white}\multicolumn{7}{c}{\textbf{Finetuned SAM2 Variants (Video-based, Sparse Supervision with Memory)}} \\
\midrule
SAM2-FT-Point-MD      & 75.65 & 80.77 & 34.21 & 37.50 & 61.86 & 70.58 \\
SAM2-FT-Point-MD+PE    & 74.48 & 79.43 & 34.57 & 37.80 & 61.51 & 70.20 \\
SAM2-FT-Bbox-MD         & 73.64 & 78.85 & 33.10 & 35.68 & 74.47 & 81.70 \\
SAM2-FT-Bbox-MD+PE      & 73.97 & 79.17 & 34.15 & 37.14 & 74.47 & 81.70 \\
SAM2-FT-Mask-MD         & 79.88 & 84.54 & 34.74 & 37.11 & 80.45 & 85.90 \\
SAM2-FT-Mask-MD+PE      & 80.06 & 84.69 & 35.20 & 37.85 & 80.43 & 85.89 \\
\bottomrule
\end{tabular}
}
\label{tab:sam2_finetuned_results_v3}
\end{table*}

\subsection{Sparse Finetuning in Surgical Video}

{Table~\ref{tab:sam2_finetuned_results_v3} presents a detailed analysis of sparse finetuning strategies with SAM2, when surgical data only contains sparse annotations (e.g., 1 out of 5 video frames are annotated). Such experiments offer a clear guideline for potential users. We evaluate two distinct sparse approaches. The upper section (Image-based, Sparse Annotation) finetunes the model using only the annotated frames from a video (with a stride of 4), discarding the intermediate frames. This represents a highly efficient, data-light approach. The lower section (Video-based, Sparse Supervision with Memory) processes all video frames, leveraging temporal information via its memory mechanism, but calculates the training loss only on the sparsely annotated frames.}

{Our results provide several practical guidelines for users. First, the more straightforward image-based finetuning (upper section) consistently achieves strong performance, especially with richer prompts like bounding boxes and masks. For users seeking a balance of high performance and low computational overhead, finetuning only the Mask Decoder (MD) stands out as a robust and efficient strategy. Second, while the video-based approach with memory is designed to harness temporal context, its effectiveness is not universal. For instance, it shows competitive performance on EndoVis2017 but struggles significantly on EndoVis2018 compared to the simpler image-based method. This performance drop can be attributed to the highly dynamic scenes in EndoVis2018, where rapid instrument and camera movements provide misleading temporal context that conflicts with the memory module's assumption of temporal consistency. This suggests that leveraging temporal data adds complexity that does not always translate to better performance, particularly when the surgical video is not smooth. Therefore, we recommend users start with the computationally cheaper image-based sparse finetuning, and only explore video-based methods if performance on their specific task plateaus and the video data is known to be relatively stable.}

\section{Discussion}\label{sec3}

We introduce SAM2, a foundation model powered by deep learning designed for the segmentation of a wide array of vision structures {for} diverse surgical video analysis. Its various prompt configurations make SAM2 a versatile tool for universal surgical image segmentation and object tracking.

Through comprehensive evaluations that include both labeled and unlabeled validation, SAM2 has demonstrated substantial capabilities in segmenting a diverse array of targets and robust generalization abilities to manage diverse data and tasks. Its performance not only significantly exceeds that of the state-of-the-art segmentation foundation model but also rivals or even surpasses specialist models in surgical image analysis. By providing a precise delineation of anatomical structures and instrument regions, SAM2 facilitates the computation of various new-appearing object measures. For example, in the field of surgical instrument tracking, SAM2 could play a crucial role in accelerating the instrument annotation process, enabling subsequent calculations of instrument 3D reconstruction, which is a critical subtask for the entire surgery flow analysis and route planning. Additionally, SAM2 provides a successful paradigm for adapting natural image foundation models to new domains, which can be further extended with fine-tuning.

Although SAM2 has strong capabilities, it shows some limitations. Except for various prompting strategies, we made ablation experiment studies and analyzed failure cases, as shown in Supplementary Section 3 and 4. One such limitation is the imbalance of recognition ability in different surgery situations. This could potentially impact the model’s performance on bad lighting surgery videos, such as Endoscapes2023. Another limitation is its difficulty in segmenting vessel-like branching structures because the bounding box prompt can be ambiguous in this setting. For example, arteries and veins share the same bounding box in endoscopy surgical images. However, these limitations do not diminish the utility of SAM2. Since SAM2 has learned rich and representative video features from the large-scale training set, it can be fine-tuned to effectively segment new objects from less-represented videos or intricate structures like ducts and vessels.

In conclusion, this study highlights the feasibility of evaluating a single foundation model capable of managing a multitude of segmentation tasks, thus eliminating the need for task-specific models. SAM2, as the first foundation model in video segmentation, has great potential to accelerate the advancement of new endoscopy surgery analysis tools and ultimately contributes to improve whole surgical flow.

\section{Methods}\label{sec4}

In this section, we introduce the details of how to deploy the SAM2 model for surgical video segmentation. Our SAM2 pipeline for surgery videos contains three steps: prompt generation, image segmentation, and video propagation. The re-configuration includes three major parts, different prompting strategies, re-initialization of prompts, and auto-segmentation. See the pipeline figure in Supplementary Figure 1.

\subsection{Prompt Extraction Strategy}

Point prompts consist of x and y coordinates, along with additional metadata. The label value (0 or 1) distinguishes between negative and positive points, allowing users to specify both object and non-object locations. Multiple point prompts can be used for a single object, providing fine-grained control over the segmentation. In our work, since our tested datasets do not include point prompts and we are not considering human labeling, we implement a comprehensive point prompt extraction strategy. This approach involves randomly sampling points from existing box or mask prompts, with additional mechanisms to introduce controlled variability. Our point extraction process involves randomly sampling N positive points inside each instance-level mask or bounding box in \(I_{p}^{0}\). For objects with multiple mask areas due to occlusion, we sample N points for each separate area. We also generate M negative points for each area, selected from positive points of other objects. To introduce variability and robustness in point selection, we implement a point fluctuation mechanism. This mechanism slightly adjusts the position of each selected point within a small radius, controlled by a hyperparameter beta. The fluctuation is applied in both x and y directions, with the new position clamped to ensure that it remains within the image boundaries.

Bounding box prompts are represented as two points, i.e., the top-left and bottom-right corners. They use special labels to distinguish them from regular point prompts: the top-left corner is labeled 2, and the bottom-right corner is labeled 3. This representation allows SAM2 to handle boxes consistent with its training in point data. While an object can only have one box prompt, it is possible to combine a box prompt with multiple additional point prompts for more precise segmentation.

Mask prompts provide pixel-level information about the object. They are binary masks where True indicates object pixels and False indicates background. The mask should have the same resolution as the input frame. Mask prompts are particularly useful for refining segmentation or propagating known segmentation to nearby frames. Typically, one object has one mask prompt.

Additionally, we consider a center point (the center of the box for bounding boxes or the center of mass for masks) as a special prompt. This strategy allows us to test the performance of SAM2 with varying levels of input information, from sparse point sets to more comprehensive representations of object boundaries.

The ability of SAM2 to handle multiframe prompts is a key feature for video segmentation tasks. Users can provide prompts on a single frame and let SAM2 propagate to other frames, or give prompts on multiple frames to guide the segmentation through challenging scenarios such as occlusions or rapid motion. This multiframe prompting capability allows for iterative refinement of segmentations by adding prompts based on initial results. Internally, SAM2 encodes these various prompt types into a unified representation that can be processed by its attention mechanisms. This allows the model to seamlessly integrate information from different prompt types and across multiple frames, enabling efficient and accurate segmentation across a wide range of video scenarios.

To process various datasets, we employ a consistent initialization strategy. Our approach handles both video clips and image datasets, supporting common formats such as jpg, png, and jpeg. For video clips, we first sample them to extract individual frames, which are then sorted in temporal order. Ground truth annotations in our tested datasets are provided in two formats: bounding boxes and masks in COCO format, or pixel-level mask images. To unify the testing process, we converted all pixel-level masks to the COCO format. From these annotations, we create a set of prompts including points, masks, and bounding boxes to test the performance of SAM2 with various input types. The first frame containing ground truth annotations is selected as the initial prompt frame, denoted as \(I_{p}^{0}\), where $p$ indicates that this frame is selected as a prompt frame. This approach mimics real-world scenarios with minimal expert annotation and tests the ability of SAM2 to propagate segmentation across videos.


\subsection{Re-initialization Strategy}\label{subsec2}
Based on preliminary results, we found that using only the first valid frame \(I_{p}^{0}\) may not be sufficient to guide a long video clip. This is because \(I_{p}^{0}\) may not contain all objects of interest, and some objects may temporarily disappear and reappear, challenging the tracking ability of SAM2. To address these issues, we implement a re-initialization strategy that can be triggered in two cases. The first case is every $T$ {frames}, where $T$ is a predefined interval. The second case occurs when a new object not present in the current prompt frame is taken into consideration. These re-initialization cases can be applied together or separately, depending on the specific requirements of the video sequence being processed. The re-initialization process involves finding a new prompt frame \(I_{p}^{t}\) and discarding the previous prompt frame, extracting object information (including identifying prompt objects, building a SAM2 video predictor, and initializing the inference state), and reinitializing the SAM2 predictor. This approach helps maintain tracking accuracy over longer video sequences and adapts to changing scene conditions. By not using previous prompts, we allow the model to focus on current information and avoid incorrect segmentation of old objects that may no longer be relevant.

\subsection{Auto-segmentation for Unannotated Datasets}\label{subsec2}
For datasets lacking ground-truth annotations, we utilize the auto mask generation techniques provided by SAM2 to automatically generate pseudo ground truth for the first frame. Our preliminary results indicate that segmentation quality is highly sensitive to certain hyperparameters. Thus, we perform an extensive hyperparameter search to find optimal settings for datasets without ground truth.

Key hyperparameters in our auto-segmentation process include the number of points sampled per side of the image, which influences the likelihood of capturing all objects, and the predicted mask quality threshold. We also consider stability score thresholds and offsets, which affect the robustness of mask generation. The non-maximal suppression threshold for filtering duplicate masks and the minimum mask region area for post-processing are crucial for refining the generated masks. Additionally, we explore the impact of multiscale processing through crop layers and the use of mask-to-mask refinement. 

The final hyperparameters for auto segmentation are chosen based on expert visual inspection, considering factors such as mask completeness, boundary accuracy, and the ability to distinguish between close or overlapping objects. This process ensures that the generated pseudo ground truth is of high quality and suitable for further processing and evaluation.

\subsection{Finetuning Strategies}\label{subsec2}

{While SAM2 is powerful out-of-the-box, its performance on rare or domain-specific tasks may not always meet expectations, especially on surgery videos. Here, we design various fine-tuning strategies to adapt SAM2 to surgery specific needs, improving the accuracy and efficiency of instrument tracking and anatomy segmentation. Our fine-tuning strategy is the process of further training the pretrained SAM2 model on the annotated surgery dataset (that are EndoVis2017, EndoVis2018, EndoNeRF, CholecSeg8k, DSAD, and Endoscapes2023). For the dataset without official training/test splitting, we split the dataset into a training set (70\%) and a testing set (30\%) to ensure that we can evaluate the model's performance after training. The training data will be used to fine-tune the SAM2 model, while the test data will be used for inference and evaluation. We finetune SAM2 by selectively training its significant modules: the mask decoder, the prompt encoder, the memory attention, the memory encoder, and the image encoder, while all other parameters remain frozen. Our image-based, per-frame finetuning strategy generates a wide range of ablation experiments by combining different sets of trainable modules with various prompting techniques to identify the best-performing setup. In addition, the sparse video-based finetuning strategy is designed and compared with sparse image-based finetuning to evaluate an efficient and scalable alternative, particularly when facing the common challenge of insufficient annotation in surgical domain.}

\subsection{Datasets}\label{subsec2}

The detailed description of the datasets is as follows: (1) EndoVis2017~\cite{endovis17} is a public dataset for laparoscopic instrument segmentation with 18 videos, split into 8 for training (1800 frames) and 10 for testing (1200 frames), featuring 15 object categories. (2)EndoVis2018~\cite{endovis18}, from the MICCAI 2018 Challenge, includes 19 videos divided into 15 for training (2235 frames) and 4 for testing (997 frames) at 1280$\times$1024 resolution, with 7 object categories. (3) EndoNeRF~\cite{endonerf} consists of two video types (``Cutting'' and ``Pulling''), showing high segmentation performance with mDice of 98.32\% and 96.09\%, and mIoU of 96.71\% and 92.87\%, respectively. (4) {DSAD~\cite{dsad} contains semantic segmentations of eight abdominal organs (colon, liver, pancreas, small intestine, spleen, stomach, ureter, vesicular glands), the abdominal wall, and two vessel structures (inferior mesenteric artery, intestinal veins) in laparoscopic view.} (5) Endoscapes2023~\cite{murali2023endoscapes}, a subset of Endoscapes2023, contains 14,940 frames from 50 videos, with 493 annotated frames covering 6 classes (anatomy and tools). (6) CholecSeg8k~\cite{hong2020cholecseg8k} includes 17 clips from Cholec80, totaling 8080 frames at 854$\times$480, with 13 categories and three mask types (color, annotation, and watershed). (7) SurgToolLoc~\cite{zia2023surgical} comprises 24,695 30-second endoscopic videos (720p, 60 fps, downsampled to 1 fps) featuring 14 surgical tools. (8) Cholec80~\cite{twinanda2016endonet} has 80 videos averaging 39 minutes at 25 fps (downsampled to 1 fps), labeled with phase and tool presence by a senior surgeon. (9) SurgicalLive24, a private diverse collection of thoracoscopy, laparoscopy, and urethroscopy videos, includes 11 operations per type, with 168,956 frames sampled at 1 fps. 

\subsection{Evaluation Metrics}\label{subsec2}

Model performance is assessed using confusion matrix components—true positives (TP), false positives (FP), and false negatives (FN)—across three primary metrics: Mean Intersection over Union (mIoU), Mean Dice Coefficient (mDice), and Mean Absolute Error (MAE). The mIoU, defined as \(\frac{1}{C} \sum_{i=1}^{C} \frac{\text{TP}_i}{\text{TP}_i + \text{FP}_i + \text{FN}_i}\), represents the average overlap between predicted and ground-truth segmentations for all classes, where \(C\) is the number of classes. The mDice, calculated as \(\frac{1}{C} \sum_{i=1}^{C} \frac{2 \times \text{TP}_i}{2 \times \text{TP}_i + \text{FP}_i + \text{FN}_i}\), quantifies the average similarity between predicted and ground-truth segmentations across all classes. For pixel-level discrepancy, MAE is employed, defined as \(\frac{1}{N} \sum_{i=1}^{N} |y_i - \hat{y}_i|\), where \(N\) is the total number of pixels, \(y_i\) is the ground-truth value, and \(\hat{y}_i\) is the predicted value for pixel \(i\).
For class-specific analysis, per-class IoU and Dice scores can be derived, enabling detailed evaluation of segmentation performance. An auxiliary overlap metric, \(\phi = \frac{\text{TP}}{\text{TP} + \text{FP} + \text{FN}}\), provides a simplified measure of segmentation consistency. This multi-faceted framework integrates overlap accuracy (mIoU), similarity assessment (mDice), and pixel-wise error (MAE) to comprehensively evaluate model efficacy.
\subsection{Experimental setting}
Unlike traditional supervised methods requiring task-specific training with manual annotations, SAM2 leverages its multi-prompt zero-shot inference framework to segment objects dynamically without retraining for new categories. The workflow involves two stages: (1) initializing targets via user-defined prompts (e.g., points, boxes, or text) on the first frame, followed by (2) automated mask propagation across subsequent frames using SAM2’s built-in temporal tracking. To enhance robustness against occlusions or drift for long-term segmentation, we integrate a custom re-initialization mechanism that re-injects prompts at predefined intervals (e.g., every 30 frames), while the baseline SAM2 system relies solely on continuous tracking when this feature is disabled. 

Experiments were conducted on a system with an AMD EPYC 7763 CPU, running Ubuntu 22.04.4 LTS with CUDA 12.3, utilizing 1 NVIDIA A6000 GPU. The software environment included PyTorch 2.3.1 and the sam2\_hiera\_large checkpoint. Videos were sampled at 1 frame per second and processed at their original resolution.

\subsection{Compared Baseline Methods}
{To evaluate the proposed framework, we benchmark against 6 existing SAM-based segmentation methods. 
They can be broadly divided into two categories: prompt-prediction and adaptation. Surgical-DeSAM~\cite{sheng2024desam} integrates surgical instrument priors via masked cross-attention modules, with a domain-specialized architecture. SelfPromptSAM~\cite{wu2023selfpromptsam} is a hybrid approach as it first trains a linear layer on top of SAM's ViT backbone to predict a segmentation mask, and then samples prompts from the predicted mask to pass to SAM. SurgicalSAM~\cite{yue2024surgicalsam} is an adaptation approach that trains a custom prompt encoder using only the target class of each object, also finetuning SAM’s mask decoder. PerSAM~\cite{zhang2305persam} and its variant named GF-SAM~\cite{zhang2024gfsam} use feature similarity to compute prompts for SAM, which fall into the prompt-prediction category. CycleSAM~\cite{murali2024cyclesam} enforces cycle-consistent mask predictions across augmented prompts.}

\section{Data and Code Availability}
This study employed nine datasets, including eight publicly available datasets and one private dataset. The public datasets comprise Endovis2017 (\url{https://endovissub2017-roboticinstrumentsegmentation.grand-challenge.org/Data/}), Endovis2018 (\url{https://endovissub2018-roboticscenesegmentation.grand-challenge.org/Data/}), EndoNeRF (\url{https://github.com/med-air/EndoNeRF?tab=readme-ov-file}), DSAD (\url{https://springernature.figshare.com/articles/dataset/The_Dresden_Surgical_Anatomy_Dataset_for_abdominal_organ_segmentation_in_surgical_data_science/21702600?file=38494425}), Endoscapes2023 (\url{https://github.com/CAMMA-public/Endoscapes}), CholecSeg8k (\url{https://www.kaggle.com/datasets/newslab/cholecseg8k}), Cholec80 (\url{https://github.com/CAMMA-public/TF-Cholec80}), and SurgToolLoc (\url{https://surgtoolloc23.grand-challenge.org/}).

The private SurgicalLive24 dataset contains sensitive patient information. Due to privacy concerns and ethical restrictions, this dataset cannot be made publicly available at this time. The processed results and aggregated statistics derived from this dataset that support the findings of this study are included in the Supplementary Materials. Access to the raw data may be granted on a case-by-case basis for academic research purposes, subject to appropriate data use agreements and ethical approvals. Requests should be directed to yban@sjtu.edu.cn.

The training and inference script has been publicly available at \href{https://github.com/Banyutong/SurgicalSAM2}{https://github.com/Banyutong/SurgicalSAM2}.

\section{Author Contribution}
Y.T.B, C.Y, J.J, and K.Y.Y conceptualized and designed the study, encompassing experimental design and data analysis; J.J, K.Y.Y and L.W formulated the framework and developed the deep learning algorithms; K.Y, L.W, R.W, Z.M, H.P, Z.X, Y.Z, and W.S executed the experiments using the provided datasets and implementations; Y.T.B, C.Y, J.J, and K.Y.Y co-authored the manuscript, with Y.B, H.S.W, Y.J, and Q.D offering valuable supervision to the study. All authors reviewed and approved the final manuscript.

\section{Acknowledgments}
This work has been supported by the program of National Natural Science Foundation of China (No. 62503322) and by Shanghai Magnolia Funding Pujiang Program (No. 23PJ1404400).

\bibliography{mybib}

\clearpage

\renewcommand{\listfigurename}{Figure legends}
{\hypersetup{linkcolor=black}\listoffigures}

\end{document}